\pdfoutput=1

\documentclass[11pt]{article}

\usepackage[final]{acl}
\usepackage{verbatim}

\usepackage{times}
\usepackage{latexsym}

\usepackage[T1]{fontenc}

\usepackage[utf8]{inputenc}

\usepackage{microtype}

\usepackage{inconsolata}

\usepackage{times}
\usepackage{latexsym}
\usepackage{bm}
\usepackage{graphicx}
\usepackage{amsmath}
\usepackage{amsfonts}
\usepackage{booktabs}
\usepackage{multirow}
\usepackage{enumitem}
\usepackage{subcaption}
\usepackage{flushend}
\usepackage{algorithm2e}
\RestyleAlgo{ruled}
\usepackage{makecell}

\title{Towards Unified Task Embeddings Across Multiple Models: Bridging the Gap for Prompt-Based Large Language Models and Beyond}

\author{
    Xinyu Wang\textsuperscript{\rm1,2}, 
    Hainiu Xu\textsuperscript{\rm2}, 
    Lin Gui\textsuperscript{\rm2}, 
    Yulan He\textsuperscript{\rm1,2,3} \\
  \textsuperscript{1}Department of Computer Science, University of Warwick \\
  \textsuperscript{2}Department of Informatics, King's College London\\
  \textsuperscript{3}The Alan Turing Institute\\
  \texttt{Xinyu.Wang.11@warwick.ac.uk} \\
  \texttt{\{hainiu.xu, lin.1.gui, yulan.he\}@kcl.ac.uk} \\
  }

\begin{document}

\maketitle

\begin{abstract}

Task embedding, a meta-learning technique that captures task-specific information, has gained popularity, especially in areas such as multi-task learning, model editing, and interpretability.
However, it faces challenges with the emergence of prompt-guided Large Language Models (LLMs) operating in a gradient-free manner.
Existing task embedding methods rely on fine-tuned, task-specific language models, which hinders the adaptability of task embeddings across diverse models, especially prompt-based LLMs.
To hardness the potential of task embeddings in the era of LLMs, we propose a framework for unified task embeddings (\texttt{FUTE}), harmonizing task embeddings from various models, including smaller language models and LLMs with varied prompts, within a single vector space.
Such uniformity enables comparison and analysis of similarities amongst different models, broadening the scope and utility of existing task embedding methods in  multi-model scenarios, while maintaining their performance comparable to architecture-specific methods.\footnote{Code is available at \href{https://github.com/xnyuwg/fute}{https://github.com/xnyuwg/fute}.}

\end{abstract}

\section{Introduction}

\begin{figure*}[t]
	\centering 
	\centerline{\includegraphics[width=0.85\textwidth]{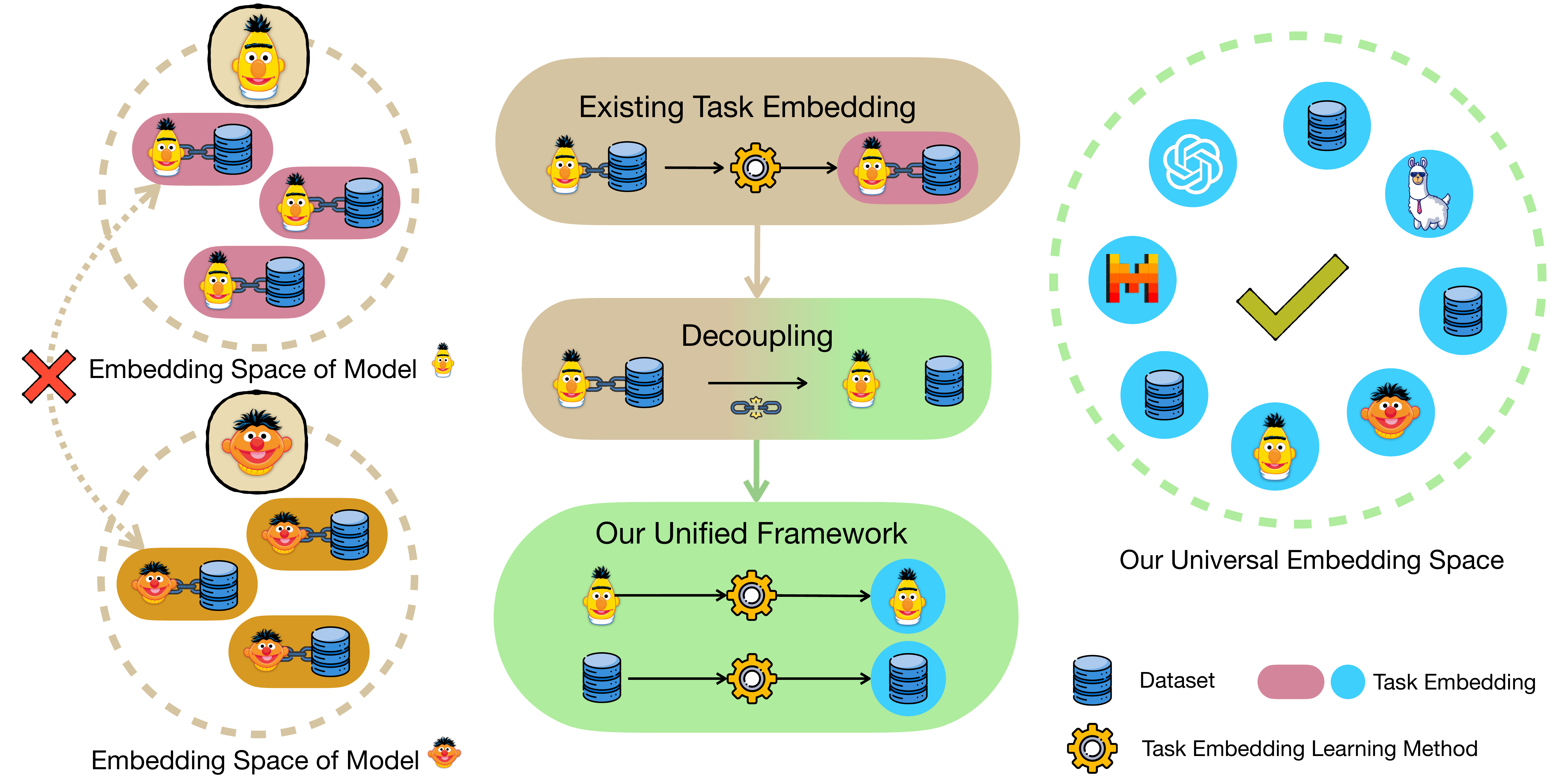}}
	\caption{Overview of \texttt{FUTE}. Existing methods generate task embeddings by leveraging model parameters, confining to models of identical architecture and initialization. This limitation, depicted on the left, results in embeddings that reside in separate vector spaces depending on the model, making cross-model similarity comparisons impossible. Our unified framework introduces a method for independently computing
 task embeddings, enabling the incorporation of diverse models, ranging from different neural architectures to Large Language Models (LLMs) with distinct prompts, into a unified vector space.}
	\label{fig:1} 
\end{figure*}

Task embeddings serve as a valuable meta-learning tool for various downstream tasks such as multi-task learning \citep{vu-etal-2020-exploring,eeml2022,zhou-etal-2022-efficiently,pmlr-v202-wu23t}, model editing \citep{ilharco2023editing,ortiz-jimenez2023task}, and interpretability \citep{gueta-etal-2023-knowledge}. 
Current task embedding methods focus on encapsulating the task-specific information into a vector representation, which is derived from parameters of task-specific, fine-tuned models \citep{vu-etal-2020-exploring,zhou-etal-2022-efficiently,ilharco2023editing}. 
However, the existing paradigm's reliance on the parameters of a single model restricts the comparison and adaptability of task embeddings across various models, especially hindering their applicability to prompt-guided, gradient-free Large Language Models (LLMs).

To overcome these challenges, we propose a model-agnostic \textbf{F}ramework for \textbf{U}nified \textbf{T}ask \textbf{E}mbedding (\textbf{\texttt{FUTE}}), which enables the application of task embedding across diverse models including language models of various architectures and LLMs with different prompts, as illustrated in Figure~\ref{fig:1}.
\texttt{FUTE} generates task embeddings by treating these models as ``black boxes'' and does not require knowledge of the original models' training datasets.
Specifically, we decouple the concept of task embedding into two distinct forms: {\emph{Dataset Task Embedding} (DTE) and {\emph{Model Task Embedding} (MTE).
DTE is designed to capture the task information specific to a dataset, focusing on its unique characteristics.
In contrast, MTE captures the model's behavior according to the task performed, independent of its training dataset.
By decoupling MTE and DTE, our framework facilitates a more granular analysis of task characteristics, providing insights into both the data and models employed. 
Moreover, the DTEs and MTEs of various models are computed using a single surrogate model, which replaces the role of the original model in existing methods. This consistent approach to computing DTEs and MTEs 
ensures that the learned task embeddings reside in the same embedding space, hence facilitating the comparison and analysis of similarities among them.

Building upon the aforementioned improvements, \texttt{FUTE} conceptualizes the combination of a prompt and a LLM as a single model unit, enabling the computation of task embeddings for LLMs guided to perform specific tasks through prompts. Experimental results show that, while being more adaptable and covering a wider range of models, our proposed framework performs on par with existing model-specific embedding methods. \texttt{FUTE} primarily focuses on extending task embedding methods to multi-model scenarios rather than directly improving the embedding methods themselves. Although the performance of FUTE relies on the employed task embedding methods, this aspect represents a strength, allowing FUTE to leverage future advancements in task embedding techniques without significant modifications.
Our main contributions are summarized as follows: 
\begin{itemize}[noitemsep,nolistsep,leftmargin=*]
\item We introduce \texttt{FUTE}, a framework capable of learning unified task embeddings from diverse models, including language models of different architectures, and LLMs with various prompts, within a single vector space.
\item We decouple the concept of task embedding into data task embedding and model task embedding. 
\item Experimental results show that \texttt{FUTE}, while being more versatile, retains a performance to be comparable to existing model-specific methods. 
\end{itemize}

\section{Related Work}

Task embedding, initially introduced as a task vector \citep{te2019}, 
as a meta-learning method for visual classification tasks. This concept utilizes the empirical Fisher information derived from model gradients to encode task-specific information into a vector representation, demonstrating that the model parameter can provide detailed task information.
Expanding into NLP, \citet{vu-etal-2020-exploring} applied Fisher information to capture task-specific details in fine-tuned language models, formalizing it as task embedding.
\citet{zhou-etal-2022-efficiently} explored three Parameter-Efficient Fine-Tuning (PEFT) methods, Prompt Tuning \citep{liu-etal-2022-p}, Bias Tuning \citep{ben-zaken-etal-2022-bitfit}, and Low-Rank Tuning \citep{hu2022lora} to generate task embeddings.
They discovered that PEFT parameters, being the only components adjusted during fine-tuning, inherently encapsulate task-specific information.
\citet{ilharco2023editing} introduced a method with arithmetic operations by subtracting the original parameters from its fine-tuned parameters to extract task vectors.
Studies by \citep{AlvarezMelis2020GeometricDD,Tan_2021_CVPR} proposed direct computation of task similarity via Wasserstein Distance, bypassing the need for embedding similarity measures.
Additional research efforts, such as \citep{lv-etal-2023-parameter,huang2023lorahub,huang-etal-2023-learning-easily} have indirectly engaged with concepts akin to task embedding for multi-task learning, albeit without explicitly naming them as such.
Recent investigations into the task embedding of LLMs by \citep{hendel-etal-2023-context} extracted task vectors from parameters and activations of LLMs. However, similar to previous approaches, this method is confined to a single model and is unable to generate task embeddings for datasets, thereby limiting its applicability.
All the aforementioned methods predominantly focused on learning task embeddings for individual models, limiting their application in scenarios involving multiple models.

\section{Methodology}

\begin{figure*}[t]
	\centering 
	\centerline{\includegraphics[width=0.999\textwidth]{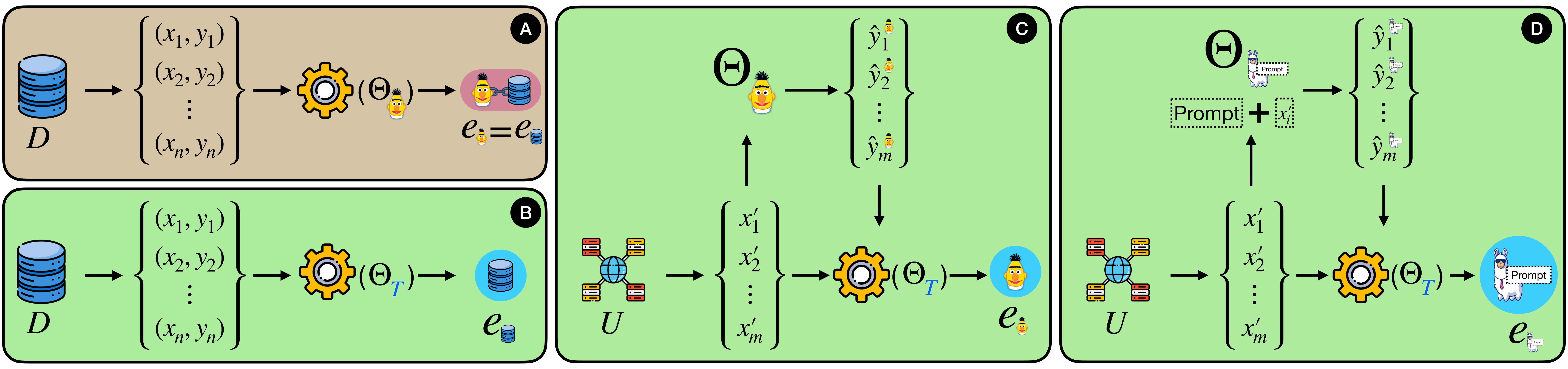}}
	\caption{Comparison between \texttt{FUTE} and existing methods. (A) Existing methods typically utilize data $D=\{(x_i, y_i)\}_{i=1}^{n}$ and model $\Theta_M$ to generate task embedding $e$ for both the dataset and the model. (B) \texttt{FUTE} derives dataset task embedding (DTE) $e_D$ by introducing an independent surrogate base model $\Theta_T$. (C) \texttt{FUTE} further advances by deriving model task embedding (MTE) $e_M$ by incorporating unsupervised data $U$ to produce alternative input $\{(x'_i, \hat{y}^M_i)\}_{i=1}^{m}$, enabling model-specific embeddings without direct dependency on task data. (D) Additionally, \texttt{FUTE} computes MTE for Large Language Models (LLMs) with prompts by treating the combination of a prompt and an LLM as a single model.}
	\label{fig:2} 
\end{figure*}

\subsection{Problem Setup}

Existing task embedding learning methods employ a singular embedding to represent both dataset and model task information. In this paper, we refine the concept  
into two distinct forms: \emph{Dataset Task Embedding} (DTE) and \emph{Model Task Embedding} (MTE). The learning of them is independent and parallel.

Formally, we define a model as $M$ and a dataset as $D=\{(x_i, y_i)\}_{i=1}^{n}$, where $x_i$ denotes the input and $y_i$ denotes the label. Task embedding is denoted as $e$, with DTE specified as $e_D$ and MTE as $e_M$. 
Typically, the task embedding learning method requires a base model, denoted as $\Theta$. The task embedding learning method $f(\cdot,\cdot)$ is formulated as:
\begin{equation}
\label{eq:f}
    e = f\big(\{(x_i, y_i)\}_{i=1}^{n}, \Theta\big) \text{,}
\end{equation}
where the function $f(\cdot,\cdot)$ employs the base model $\Theta$ and a set of data instances $\{(x_i, y_i)\}_{i=1}^{n}$ to compute the corresponding task embedding $e$.
In this paper, $f(\cdot,\cdot)$ encapsulates the complete pipeline of a task embedding learning method, integrating all steps from initial data to the final embedding output.
It is important to note that to ensure all task embeddings reside within the same vector space, it is crucial that the function $f(\cdot,\cdot)$, responsible for learning task embeddings, utilizes a consistent base model $\Theta$. 
This requirement ensures the consistency of the embedding process across different tasks.

\subsection{Background}
\label{sec:background}

In this paper, we concentrate on extending two task embedding learning methods: TaskEmb \citep{vu-etal-2020-exploring} and TuPaTE \citep{zhou-etal-2022-efficiently}. TaskEmb employs the Fisher Information Matrix (FIM), which captures the curvature of the loss surface, to learn task embeddings. TuPaTE, on the other hand, introduces Parameter-Efficient Fine-Tuning (PEFT) methods to learn task embedding.

\paragraph{TaskEmb}
\label{sec:taskemb}

TaskEmb \citep{vu-etal-2020-exploring} employs the FIM with respect to the parameters of a language model to compute task embeddings. 
The FIM is derived from the expected covariance of the gradients concerning the parameter $\theta$ of a language model $\Theta_M$ where $\theta \in \Theta_M$ as:
\begin{equation} \small
F_{\theta} = 
\displaystyle \mathop{\mathbb{E}}_{x,y \sim P_\theta(x,y)}
 \nabla_{\theta} \log P_\theta(y|x) \nabla_{\theta} \log P_\theta(y|x)^{T}\text{.}   
\end{equation}
Specifically, TaskEmb computes the empirical Fisher on a fine-tuned model $\Theta_M$ using the  
labels: 
\begin{equation} \small
F_{\theta} = \frac{1}{n}\sum\limits_{i=1}^{n} \left[ \nabla_{\theta} \log P_\theta(y_i|x_i) \nabla_{\theta} \log P_\theta(y_i|x_i)^{T}\right]\text{.}
\end{equation}
Subsequently, the diagonal entries are extracted and averaged across the entire dataset, and then concatenated to form the task embedding $e$. The complete pipeline of TaskEmb is denoted as $e=f_{\text{TaskEmb}}(\{(x_i, y_i)\}_{i=1}^{n}, \Theta_M)$.

\paragraph{TuPaTE}
\label{sec:tupate}

TuPaTE demonstrates the application of three PEFT methods: Prompt Tuning \citep{liu-etal-2022-p}, Bias Tuning \citep{ben-zaken-etal-2022-bitfit}, and Low-Rank Tuning \citep{hu2022lora}. In this paper, our focus is specifically on TuPaTE with Prompt Tuning. However, \texttt{FUTE} can be easily adapted to other methods in a similar manner.

Specifically, with a language model $\Theta_M$, TuPaTE with Prompt Tuning incorporates additional attention prefix matrices $K_p$ and $V_p$ as parameters.
For each layer, these augmented vectors are concatenated with the existing key $K$ and value $V$ 
to form $K'=[K_p,K]$ and $V'=[V_p,V]$, where $[\cdot,\cdot]$ denotes concatenation. Subsequently, the attention mechanism employs the modified $K'$ and $V'$ for computation.
Following PEFT using the prompt $K_p$ and $V_p$ in the language model $\Theta_M$ on dataset $\{(x_i, y_i)\}_{i=1}^{n}$, these prompt vectors are averaged across all layers and concatenated to form the task embedding $e$. The entire process of TuPaTE is denoted as $e=f_{\text{TuPaTE}}(\{(x_i, y_i)\}_{i=1}^{n}, \Theta_M)$.

\subsection{Model Task Embedding}
\label{sec:mte}

Model Task Embedding (MTE) is designed to encapsulate the task-specific capabilities of a model within an embedding. 
Previous methods require that task embeddings be derived from the same base model $\Theta_M$. 
Furthermore, these 
methods are closely tied to datasets, constraining the embeddings to models trained on specific datasets, as illustrated in Figure~\ref{fig:2}A. 
The dependency on datasets 
restricts applicability in scenarios where datasets, such as those for LLMs, are unavailable or inaccessible due to privacy concerns.
To overcome these limitations, we introduce a model-agnostic and dataset-agnostic approach for extracting MTE. 
This method offers 
advantages in scenarios involving varied models and in contexts where training data is inaccessible.

\paragraph{MTE Framework}

\texttt{FUTE} eliminates the dependency on a specific dataset $D$ for understanding the task information of a model, as depicted in Figure~\ref{fig:2}C,
by utilizing an unsupervised text data $U=\{x'_i\}_{i=1}^{m}$,  
unrelated to specific tasks, to substitute the conventional task-specific dataset $D=\{(x_i, y_i)\}_{i=1}^{n}$.
By applying the model $\Theta_M$ on $U=\{x'_i\}_{i=1}^{m}$, we first generate predictions $\{\hat{y}^M_i\}_{i=1}^{m}$ according to:
\begin{equation} \small
    \hat{y}^M_i = \Theta_M (x'_i) \text{.}
    \label{eq:MTE-prediction}
\end{equation}
This approach is based on the premise that different task-specific models,  
such as sentiment analysis or question answering, will yield divergent outcomes even for identical text inputs.
Here, $\Theta_M (x'_i)$ captures task-related information inherent to  
$\Theta_M$.

Additionally, \texttt{FUTE} introduces a surrogate base model for task embedding learning, denoted as $\Theta_T$, as a substitute for $\Theta_M$.
This substitution preserves the uniformity of $\Theta_T$ throughout the task embedding learning process, eliminating the need for the same $\Theta_M$. 
The T5-base \citep{2020t5} is employed as $\Theta_T$ in our experiments.
We pair each $x'_i$ with its corresponding task-related $\hat{y}^M_i$ to feed into the task embedding learning method $f(\cdot,\cdot)$, deriving MTE $e_M$ as:
\begin{equation} \small
\label{eq:em}
    e_M = f(\{(x'_i, \hat{y}^M_i)\}_{i=1}^{m}, \Theta_T) \text{.}
\end{equation}
This mechanism allows for the adaptation of the MTE learning process to various models, with the model output $\{\hat{y}^M_i\}_{i=1}^{m}$ reflecting the nature of the task of $\Theta_M$, be it a probability, class, or sentence.

\paragraph{MTE for Prompt-based LLMs}

Previous  
methods typically rely on task-specific model parameters and datasets, which are not applicable or accessible for LLMs, especially those closed-source LLMs. Conversely, \texttt{FUTE} can be easily extended to accommodate prompt-based LLMs. In essence, a prompt in conjunction with an LLM can be treated as a unified model from which its MTE can be derived. Prompts can vary from simple instructions to more complex In-Context Learning examples (ICL) and Chain of Thought (CoT) prompts. 

We use the same MTE framework proposed above, and the only difference is that the input to Eq.~(\ref{eq:MTE-prediction})  includes a prompt in addition to the original input $x'$, as shown in Figure~\ref{fig:2}D. Even though the LLM here, $\Theta_M$, may be significantly larger in size compared to the MTE learning model, $\Theta_T$, 
an LLM guided by a 
prompt, which is limited to performing a specific task, can be managed by $\Theta_T$.

This method holds particular promise for LLMs, 
where different prompts can trigger varied task-specific performances, thereby offering a nuanced understanding of the task capability of LLMs without being limited to specific fine-tuned parameters.

\paragraph{Sourcing Unsupervised Data}

For sourcing unsupervised text data $U$, we opt for the CrossFit dataset \citep{ye-etal-2021-crossfit}. CrossFit encompasses a dataset collection of 160 diverse NLP tasks, offering a rich variety of text styles and structures. 
This diversity stands in contrast to large, unsupervised text corpora such as Wikipedia, which predominantly feature formal and neutral tones, and consistent style.
Such uniformity in text data can skew the performance of models towards, for example, neutral predictions in tasks such as sentiment analysis. This bias may require a much larger volume of data to develop a comprehensive understanding of the model's capabilities.
Given this context, the broad spectrum of text characteristics found in CrossFit makes it an ideal choice for $U$ in our study, especially when considering the efficiency of MTE learning.

\paragraph{Discussion of MTE}

Viewed from an alternative perspective, this MTE learning approach resembles the principles of knowledge distillation, as described in \citep{Hinton2015DistillingTK}, by utilizing extensive unsupervised data. It effectively facilitates the transfer of knowledge from the original model, $\Theta_M$, to the surrogate base model, $\Theta_T$.

Theoretically, \texttt{FUTE} can learn MTEs for any models with accessible outputs $\{\hat{y}^M_i\}_{i=1}^{m}$, which potentially encompass not only language models and LLMs but also traditional neural networks like CNN \citep{Fukushima1980NeocognitronAS} and LSTM \citep{Hochreiter1997LongSM}, as well as classical models such as decision trees and Support Vector Machines \citep{svm1998}.

\subsection{Dataset Task Embedding}

Dataset Task Embedding (DTE) aims to encapsulate the task information of a dataset within an embedding. 
In this paper, DTE is conceptualized similarly to the task embedding methods presented in prior studies. 
The process of learning DTE closely aligns with that of MTE utilizing $\Theta_T$.
Specifically, as illustrated in Figure~\ref{fig:2}B, given a dataset $D=\{(x_i, y_i)\}_{i=1}^{n}$, the computation of DTE $e_D$ is revised as follows:
\begin{equation} \small
\label{eq:ed}
    e_D = f\big(\{(x_i, y_i)\}_{i=1}^{n}, \Theta_T\big) \text{.}
\end{equation}
This reformulation facilitates a more flexible and adaptable approach for generating task embeddings, by leveraging $\Theta_T$ without being bounded by the constraints of architectural and initialization uniformity imposed by $\Theta_M$.

\section{Experiments}

We conducted two sets of experiments focused on smaller language models and LLMs.

\subsection{Language Model Experiments}
\label{sec:lmexp}

\begin{table*}[t]
\centering
\resizebox{0.7\textwidth}{!}{
\begin{tabular}{lcccccccc}
\toprule
\multirow{4}{*}{Method} & \multicolumn{4}{c}{\textbf{CR}} &
\multicolumn{4}{c}{\textbf{QA}} \\
\cmidrule(l){2-5} \cmidrule(l){6-9}
& \multicolumn{2}{c}{\emph{in-class}} & \multicolumn{2}{c}{\emph{all-class}} & \multicolumn{2}{c}{\emph{in-class}} & \multicolumn{2}{c}{\emph{all-class}} \\
\cmidrule(lr){2-3} \cmidrule(lr){4-5} \cmidrule(lr){6-7} \cmidrule(lr){8-9} 
& $\rho\downarrow$ & NDCG $\uparrow$ & $\rho\downarrow$ & NDCG $\uparrow$ & $\rho\downarrow$ & NDCG $\uparrow$ & $\rho\downarrow$ & NDCG $\uparrow$ \\
\midrule 
DataSize & 3.6	& 80.4	& 7.8	& 75.2	& 3.2	& 84.4	& 11.4	& 65.8 \\
CurveGrad & 5.5	& 68.6 & - & - & 8.3 & 64.8 & - & - \\
TextEmb	& 5.2	& 76.4	& 9.8	& 74.7 & 4.1	& 81.1	& 5.8	& 82.0 \\
TaskEmb	& 2.8 & 82.3	& 5.4	& 78.3 & 3.2	& 84.5	& 5.4	& 82.8 \\
TuPaTE	& \textbf{2.5}	& 83.7	&  \textbf{4.5} &   81.0 & \textbf{3.0}	& \textbf{85.7}	& 4.8	& 83.3 \\
\midrule 
\texttt{FUTE} + TaskEmb	& 4.4	& 79.4	&  7.0 & 77.9 & 4.5	& 83.5	& 5.3	& 84.3 \\
\texttt{FUTE} + TuPaTE & 3.3	& \textbf{83.8}	&  6.2 & \textbf{82.0} & 3.3	& 85.6	& \textbf{4.1}	& \textbf{84.8} \\
\bottomrule
\end{tabular}
}
\caption{Transferability results of average rank $\rho$ and NDCG (\%) on classification or regression (CR) and question answering (QA) datasets. The terms \emph{in-class} and \emph{all-class} denote scenarios where the source datasets are within the same category (CR or QA) or across the two categories, respectively. For example, \emph{in-class} results for CR indicate that the source datasets are exclusively from the CR category.}
\label{table:trans}
\end{table*}

\subsubsection{Experimental Setup}

To assess the effectiveness of \texttt{FUTE} with  
language models, we follow \citep{vu-etal-2020-exploring, zhou-etal-2022-efficiently} to evaluate the transferability  
across 11 Classification or Regression (CR) and 11 Question Answering (QA) datasets.
\citet{vu-etal-2020-exploring, zhou-etal-2022-efficiently} conducted a transfer learning experiment, where initially, a language model is fine-tuned on a source dataset, followed by further fine-tuning on a target dataset.
Based on the similarity between task embeddings learned from BERT-base \citep{devlin-etal-2019-bert}, a source dataset is selected for a given target dataset, promising to achieve optimal transfer learning gain.
Evaluating the selection process  
highlights the effectiveness of task embedding. 

Our approach to the experiment diverges from the original method by shifting the focus from selecting the most appropriate source dataset to identifying the most compatible models for a target dataset.
This is achieved by matching the MTEs of candidate models with the DTE of the target dataset based on their embedding similarity.
More experimental setup details are given below:

\paragraph{Datasets}

CR datasets comprise nine datasets from the GLUE benchmark \cite{wang2018glue}, in addition to SNLI \citep{bowman-etal-2015-large} and SciTail \citep{Khot_Sabharwal_Clark_2018}. QA datasets contain eleven datasets from the MultiQA repository \citep{MultiQA19}. See Appendix~\ref{app:dataset_lm} for more details.

\paragraph{Unsupervised Dataset}

We randomly sampled 100 text entries from each task within CrossFit \citep{ye-etal-2021-crossfit} 
to ensure broad text styles and structures. We removed any datasets that overlap with the CR or QA datasets mentioned above, 
discarding any associated labels. This process resulted in the compilation of 104,162 text entries for $U$. In our experimental setup, we fine-tune $\Theta_T$ on the unsupervised data for one epoch.

\paragraph{Metric}

We follow \citep{vu-etal-2020-exploring, zhou-etal-2022-efficiently} and employ two metrics: (1) the average rank $\rho$ of the candidate source model that yields the highest transfer gain; (2) the Normalized Discounted Cumulative Gain (NDCG) \citep{ndcg2002}, which assesses the overall quality of the entire ranking. 

We present two extensions for TaskEmb and TuPaTE with \texttt{FUTE}:
(1)~\textbf{\texttt{FUTE}+TaskEmb}: It combines \texttt{FUTE} with the TaskEmb method, utilizing Eq.~(\ref{eq:em}) 
and (\ref{eq:ed}), 
in conjunction with the TaskEmb method $f_{\text{TaskEmb}}(\cdot,\cdot)$. 
(2)~\textbf{\texttt{FUTE}+TuPaTE}: It applies \texttt{FUTE} on TuPaTE, employing Eq.~(\ref{eq:em}) and (\ref{eq:ed}) alongside TuPaTE method $f_{\text{TuPaTE}}(\cdot,\cdot)$.

\subsubsection{Baselines}

In language model experiment, we compare \texttt{FUTE} with the following baselines:
(1) \textbf{DataSize} \citep{vu-etal-2020-exploring} prioritises dataset based on the size of dataset.
(2) \textbf{CurveGrad} \citep{bingel-sogaard-2017-identifying, vu-etal-2020-exploring} analyzes the gradients of the loss curve associated with the language model, 
and is adapted by \citep{vu-etal-2020-exploring} for transferability prediction.
(3) \textbf{TextEmb} \citep{vu-etal-2020-exploring} computes an average of the text representations extracted from the last layer of the language model across the entire dataset.
(4) \textbf{TaskEmb} \citep{vu-etal-2020-exploring}, as detailed in Section~\ref{sec:taskemb}, leverages Finisher information to derive task embedding.
(5) \textbf{TuPaTE} \citep{zhou-etal-2022-efficiently}, outlined in Section~\ref{sec:tupate}, utilize PEFT with the language model, extracting the tuned prompt as task embedding.

All these baselines are unavailable in the multi-model scenario. 
Dataset-centric methods, like DataSize, inherently cannot differentiate between models trained on the same dataset. 
Model-specific approaches, CurveGrad, TextEmb, TaskEmb, and TuPaTE, all require a single model initialization, limiting their adaptability across diverse models.
CurveGrad computes the gradients from the same model.
TextEmb extracts the text representations from the same model. 
For example, similarly, text embedding derived from BERT-base, BERT-large, and GPT2 do not reside in the same vector space. 
However, \texttt{FUTE} can be potentially adapted to extend text embedding methods as well.
TaskEmb and TuPaTE, as mentioned above, are also unavailable for multiple models.

\subsubsection{Transferability Results}
\label{sec:trans}

The Transferability results are 
presented in Table~\ref{table:trans}. 
Upon comparison, \texttt{FUTE} exhibits competitive performance relative to the original methods.
Specifically, when integrated with TaskEmb, \texttt{FUTE} achieves better results in the QA \emph{all-class} setting compared to the original TaskEmb method, albeit with slightly less optimised outcomes in other settings.
Similarly, the combination of \texttt{FUTE} with TuPaTE demonstrates improved performance in the \emph{all-class} setting, while yielding very similar outcomes in the \emph{in-class} setting.
Overall, the discrepancies between \texttt{FUTE} and the original methods are generally confined to less than a 2\% difference.

\subsubsection{Cross-Domain Results}

\texttt{FUTE} utilizes consistent unsupervised data $U$ for MTEs across all models.
This approach diverges from previous methodologies, where each task embedding is 
tied to a specific dataset.
We hypothesise that task embeddings generated by previous methods may encapsulate domain-specific information due to their dataset-centric nature. 
To test this hypothesis, we conduct experiments using three datasets from the finance and medical domains: FinancialPhraseBank \citep{Malo2014GoodDO}, Medical Question Pairs (MQP) \citep{mccreery2020effective}, and MedQA \citep{jin2021disease}. 
We randomly split 20\% of the training set to serve as the validation set.
These datasets exhibit text characteristics distinct from the general text content presented in the datasets used in Section~\ref{sec:trans}.

The comparative results, displayed in Table~\ref{table:cross}, reveal that TaskEmb maintains consistent performance across this cross-domain experiment, closely mirroring the results obtained from \texttt{FUTE}.
TaskEmb leverages the parameters of the entire language model, potentially embedding within it a wider range of domain information. 
Conversely, TuPaTE 
extracts task embeddings from the PEFT parameters, which may be more narrowly tailored to specific domain characteristics.
This difference suggests that TaskEmb's embeddings could 
carry broader 
information due to its reliance on the comprehensive model parameters.

\begin{table}[h]
\centering
\resizebox{0.3\textwidth}{!}{
\begin{tabular}{lcc}
\toprule
Method & $\rho\downarrow$ & NDCG $\uparrow$ \\
\midrule 
TaskEmb	& 5.6	& 80.9 \\
TuPaTE	& 10.7	& 73.8 \\
\cmidrule(lr){1-3}
\texttt{FUTE} + TaskEmb	& \textbf{4.7}	& \textbf{81.8} \\
\texttt{FUTE} + TuPaTE	& 7.0	& 81.1 \\
\bottomrule
\end{tabular}
}
\caption{Cross-domain results of average rank $\rho$ and NDCG (\%) on financial and medical domain datasets.}
\label{table:cross}
\end{table}

Remarkably, \texttt{FUTE} using TuPaTE significantly outperforms the original method.
This improvement can be attributed to the utilization of consistent unsupervised data $U$ across all models, thereby reducing the domain influence from 
the data source.
The findings from Table~\ref{table:cross} indicate the promising potential of \texttt{FUTE} to enhance the adaptability and effectiveness 
in cross-domain applications.

\subsubsection{Visualization}

In this section, we present a visualization in Figure~\ref{fig:3} by extending TuPaTE to three language models, BERT-base \citep{devlin-etal-2019-bert}, GPT2 \citep{radford2019gpt2}, and T5-base \citep{2020t5}, based on with CR and QA datasets.
A notable observation from the visualization is the emergence of two main clusters based on task information, highlighting the 
significant distinctions of MTEs between task categories. 
MTEs associated with the same language model, particularly for QA tasks, demonstrate a tendency to cluster together,
which indicates that the inherent characteristics of a language model also influence the task embedding space.
An interesting detail is the presence of three blue points within the CR task cluster, representing MTEs associated with the BoolQ dataset \citep{clark-etal-2019-boolq}. 
This positioning matches BoolQ's characteristics with CR tasks due to its requirement for boolean answer generation for the question, which closely resembles CR tasks. 
This positioning matches with the characteristics of BoolQ, which is formatted to predict a boolean answer for each question and closely resembles CR tasks.
See Appendix~\ref{app:vis_lm} for more details.

\begin{figure}[h]
	\centering 
	\centerline{\includegraphics[width=0.499\textwidth]{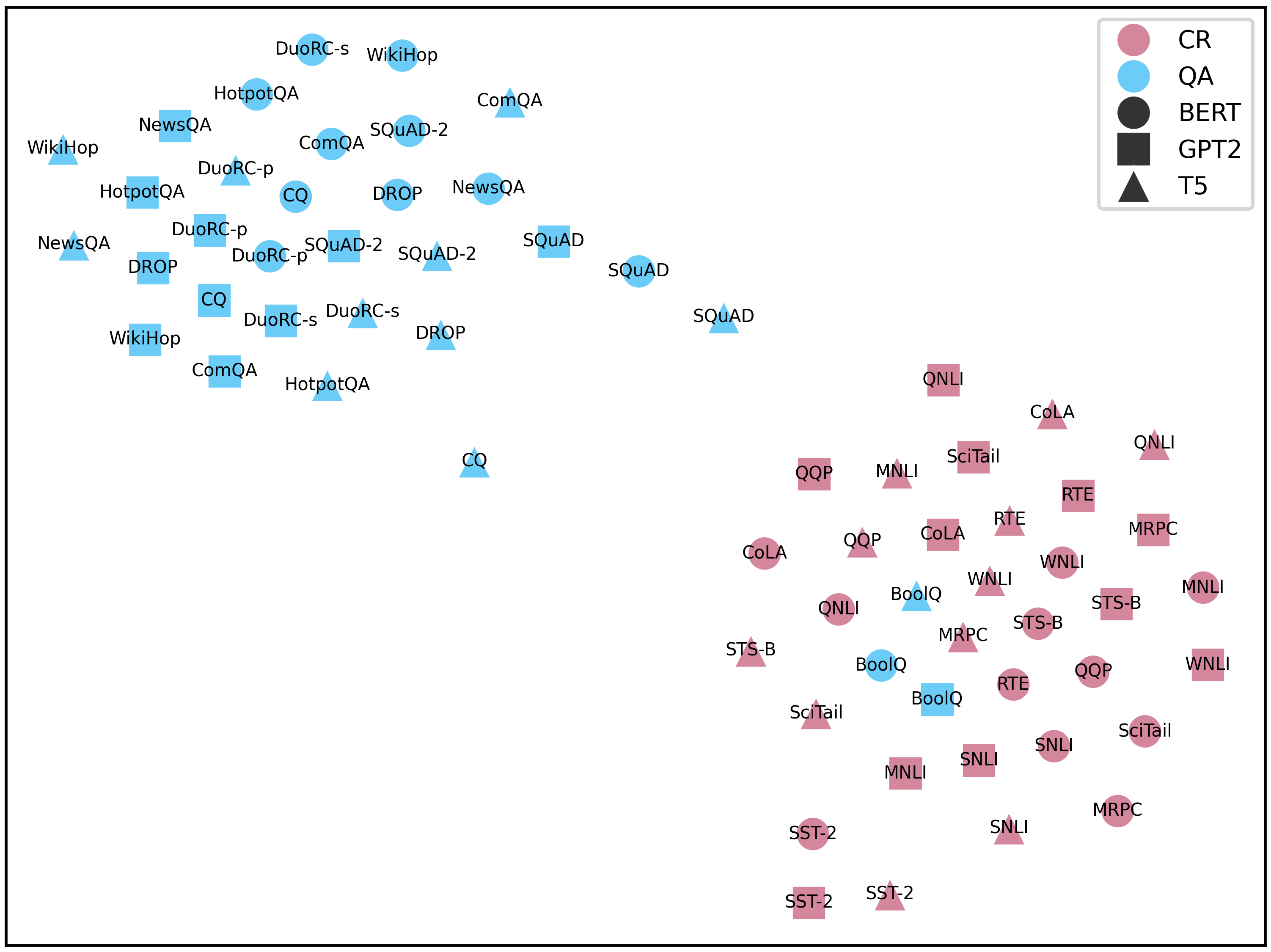}}
	\caption{T-SNE visualization of language model MTEs using \texttt{FUTE} with TuPaTE. Different task categories are represented by various colors, while different models are indicated by distinct shapes. E.g., a red circle highlights an MTE of BERT trained on a CR 
 dataset.}
	\label{fig:3} 
\end{figure}

\subsection{LLMs Experiments}
\label{sec:llmexp}

\begin{table*}[t!]
\centering
\resizebox{0.9\textwidth}{!}{
\begin{tabular}{clccccccccc}
\toprule
\multirow{4}{*}{Category} & \multirow{4}{*}{Method} & \multicolumn{3}{c}{Llama 2 13B} & \multicolumn{3}{c}{Llama 2 70B} & \multicolumn{3}{c}{Mixtral 8x7B} \\
\cmidrule(lr){3-5} \cmidrule(lr){6-8} \cmidrule(lr){9-11}
& & Performance & Rate & NDCG & Performance & Rate & NDCG & Performance & Rate & NDCG \\
\midrule 
\multirow{11}{*}{\textbf{SA}}
& MI & 88.0	& 94.4	& 59.0 & 85.3 & 90.9	& 72.9 & 87.2	& 85.1	& 65.1 \\
& LocalE & 84.3	& 90.2	& 47.5 & 88.3 & 88.7	& 57.5 & \textbf{88.5}	& \textbf{96.7}	& 78.6 \\
& GlobalE & 89.2	& 95.7	& 88.8 & 91.9 & 97.9	& \textbf{82.7} & 88.4	& 96.6	& \textbf{86.7} \\
& ZPS-Log & 54.3	& 58.2	& 38.5 & 78.0 & 83.0	& 54.0 & 57.0	& 62.1	& 33.9 \\
& ZPS-Prob & 54.3	& 58.2	& 38.5 & 78.0 & 83.0	& 50.8 & 57.0	& 62.1	& 33.9 \\
& ZPS-Vote & 54.3	& 58.2	& 38.5 & 78.0 & 83.0	& 50.8 & 57.0	& 52.1	& 33.9 \\
& Self-Select & 54.3	& 58.2	& 42.8 & 85.3 & 90.9	& 69.3 & 57.0	& 62.1	& 38.5 \\
& SPELL & 89.2	& 95.7	& \textbf{89.6} & 79.6 & 84.6	& 65.4 & 57.0	& 62.1	& 38.2 \\
\cmidrule(lr){2-11}
& \texttt{FUTE} + TaskEmb & \textbf{89.6}	& \textbf{96.1}	& 89.4 & \textbf{93.0} & \textbf{99.0}	& 74.5 & 86.9	& 94.9	& 71.1 \\
& \texttt{FUTE} + TuPaTE & 89.2	& 95.7	& 55.6 & 92.4 & 98.4	& 67.9 & 87.9	& 95.8	& 52.2 \\
\midrule 
\multirow{11}{*}{\textbf{NLI}}
& MI & \textbf{46.8}	& \textbf{90.1}	& 61.7 & 48.6 & 94.8	& 74.6 & 37.2	& 73.6	& 36.7 \\
& LocalE & 37.5	& 74.4	& 56.4 & 43.9 & 84.6	& 66.6 & 39.1	& 78.6	& 43.5 \\
& GlobalE & 40.4	& 80.4	& 65.3 & 48.2 & 93.7	& 76.9 & 40.2	& 79.1	& 44.1 \\
& ZPS-Log & 34.8	& 70.2	& 48.1 & 34.9 & 66.8	& 49.5 & 39.0	& 76.5	& 41.7 \\
& ZPS-Prob & 32.6	& 65.6	& 39.7 & 38.0 & 73.2	& 51.2 & 39.5	& 78.9	& 39.3 \\
& ZPS-Vote & 32.6	& 65.6	& 39.7 & 33.7 & 64.3	& 48.4 & 39.5	& 78.9	& 39.3 \\
& Self-Select & 33.9	& 68.1	& 39.5 & 39.7 & 76.7	& 53.6 & 39.1	& 78.6	& 43.9 \\
& SPELL & 42.4	& 84.4	& \textbf{78.1} & 48.6 & 94.8	& 77.2 & 39.1	& 78.6	& 41.5 \\
\cmidrule(lr){2-11}
& \texttt{FUTE} + TaskEmb & 35.8	& 72.0	& 47.4 & 41.1 & 78.8	& 60.8 & \textbf{43.2}	& \textbf{85.6}	& \textbf{49.1} \\
& \texttt{FUTE} + TuPaTE & 37.0	& 75.0	& 71.8 & \textbf{50.6} & \textbf{98.4}	& \textbf{81.8} & 40.8	& 81.3	& 44.4 \\
\bottomrule
\end{tabular}
}
\caption{Zero-shot prompt selection results of Performance (\%), Performance Rate (\%), and NDCG (\%) on Sentiment Analysis (SA) and Natural Language Inference (NLI) datasets.}
\label{table:llm}
\end{table*}

\subsubsection{Experimental Setup}

In this section, we assess the effectiveness of \texttt{FUTE} for LLMs by focusing on the task of zero-shot prompt instruction selection.
The objective is to identify the most suitable prompt instruction that enables an LLM to achieve optimal performance on a given dataset.
We utilize \texttt{FUTE} to generate MTEs of LLMs associated with prompts using Eq. (\ref{eq:em}), and compute DTE for the target dataset using Eq. (\ref{eq:ed}).
We then compute the similarity between MTEs and DTEs for prompt selection.

\paragraph{Datasets}

We utilize two categories of tasks for evaluation: Sentiment Analysis (SA) and Natural Language Inference (NLI). 
For SA, we employ: Rotten Tomatoes \citep{pang-lee-2005-seeing}, IMDB \citep{maas-etal-2011-learning}, and SST-2 \citep{socher-etal-2013-recursive}.
For NLI, we use: WANLI \citep{liu-etal-2022-wanli}, CB \citep{de2019commitmentbank}, and ANLI (Round 3) \citep{nie-etal-2020-adversarial}. We randomly sample 1,000 data from the test set of each dataset for evaluation.

\paragraph{Unsupervised Dataset}

We employ CrossFit \citep{ye-etal-2021-crossfit} as the source of unsupervised data $U$ and randomly sample 10,000 diverse text entries from $U$ in Section~\ref{sec:lmexp}. 
Again, we ensure no overlap between $U$ and datasets $D$, and exclusively use text content, discarding any associated labels.

\paragraph{LLMs}

We employ Llama2-13B \citep{Touvron2023Llama2O} in 8 bit, Llama2-70B \citep{Touvron2023Llama2O} in 4 bit, and Mixtral 8x7B \citep{Jiang2024MixtralOE} in 8 bit.
Mixtral 8x7B is a Sparse Mixture of Experts (SMoE) architecture, featuring 8 experts with each token processed by two experts. 

\paragraph{Prompts}

We incorporate a diverse set of 13 zero-shot prompts for each category of tasks, which comprises 10 vanilla instruction prompts sourced from PromptSource \citep{bach2022promptsource} and 3 Chain of Thought (CoT) prompts. Each prompt is structured to include instructions only, without any ICL examples.
See Appendix~\ref{app:instructions} for more details.

\paragraph{Metric}

We utilize three metrics: (1) the performance of the selected prompt; (2) performance rate, computed as the ratio of the achieved performance with the selected prompt to the maximum performance observed across all prompts. A performance rate of 100\% indicates perfect selection; (3) NDCG \citep{ndcg2002} evaluates the overall quality of prompt ranking, where relevance is determined by performance. An NDCG of 100\% represents an ideal ranking. The final results are averaged across datasets.

\subsubsection{Baselines}

We compare \texttt{FUTE} against the following baselines:
(1) \textbf{MI} \citep{sorensen-etal-2022-information} selects the prompt that 
maximizes the mutual information between the input and the LLM output.
(2) \textbf{LocalE} \citep{lu-etal-2022-fantastically} calculates the local entropy of the prediction probabilities for each entry.
(3) \textbf{GlobalE} \citep{lu-etal-2022-fantastically} computes the global entropy across all prediction categories in a dataset.
\textbf{ZPS} \citep{Liao2022ZeroLabelPS} leverages an ensemble of prompts to generate pseudo labels and then predicts the performance of each prompt on these pseudo labels. Based on the ensembling technique, it has three variants: (4) \textbf{ZPS-Log} using the sum of log probabilities, (5) \textbf{ZPS-Prob} using the sum of probabilities, and (6) \textbf{ZPS-Vote} using majority vote.
(7) \textbf{Self-Select} \citep{ramji2023selfselect} introduces a meta-prompt that queries LLM 
for selecting the most effective 
prompt.
(8) \textbf{SPELL} \citep{gonen-etal-2023-demystifying} predicts prompt performance based on the perplexity of the prompt.

Here, we also use two extensions with \texttt{FUTE}: \textbf{\texttt{FUTE}+TaskEmb} with $f_{\text{TaskEmb}}(\cdot,\cdot)$ and \textbf{\texttt{FUTE}+TuPaTE} with $f_{\text{TuPaTE}}(\cdot,\cdot)$.

\subsubsection{Overall Results}

Table~\ref{table:llm} presents the comparative results. No single baseline consistently outperforms the others across all scenarios.
Focusing on performance and rate performance,
MI excels on the Llama2-13B for the NLI task.
LocalE stands out on the Mixtral for the SA task.
\texttt{FUTE}+TaskEmb achieves superior performance on Llama2-13B and Llama2-70B for the SA task.
However, given that performance and rate performance only focus on top-1 selection, we delve deeper into the NDCG scores for a comprehensive quality assessment.

GlobalE obtains the highest NDCG on Llama2-70B and Mixtral for the SA task.
Its methodology, based on model entropy and confidence, appears less effective for the more challenging NLI task, likely due to the increased uncertainty even for correct prediction in LLM responses.
SPELL achieves the best NDCG on Llama2-13B for both SA and NLI task, but underperforms on Llama2-70B and Mixtral.
Given SPELL's reliance on perplexity, this outcome may suggest that larger models possess greater resilience in delivering accurate predictions, even in scenarios of poor perplexity. 

\texttt{FUTE} outperforms others on Llama2-70B and Mixtral for NLI tasks, demonstrating the effectiveness of our framework when handling difficult tasks.
For the Llama2 model, \texttt{FUTE}+TaskEmb outperforms \texttt{FUTE}+TuPaTE on the SA task, but underperforms on NLI task.
This disparity may stem from the inherent complexity of the NLI task, which tends to produce more unconfident and potentially noisier predictions from LLMs.
TaskEmb leverages the parameters of the entire model, learning a broad knowledge but also risking overfitting to noisy data.
TuPaTE utilizing the PEFT parameter, has a more constrained capacity that may better isolate relevant, non-noisy information.
This specialization likely accounts for TaskEmb's superior performance in SA tasks and vice versa.
On Mixtral, which is based on SMoE architecture, \texttt{FUTE}+TaskEmb demonstrates superior results. The diverse experts within Mixtral may require more information storage capacity, a requirement that TaskEmb meets.
In summary, while each baseline excels in certain tasks/models, \texttt{FUTE} emerges as a stable approach, achieving top-3 results across various scenarios.

\subsubsection{Visualization}

\begin{figure}[h]
	\centering 
	\centerline{\includegraphics[width=0.499\textwidth]{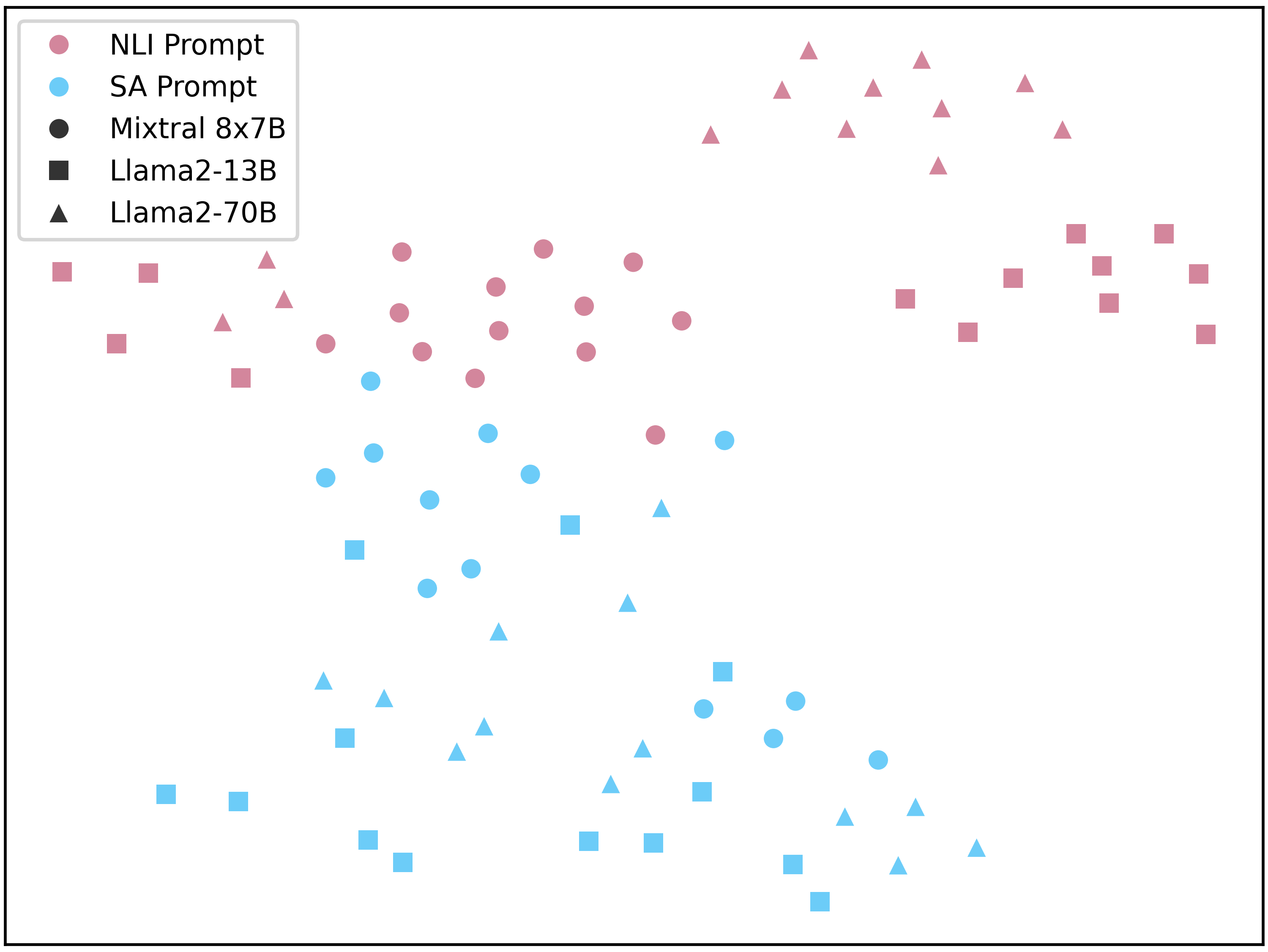}}
	\caption{T-SNE visualization of LLMs MTE using \texttt{FUTE} with TuPaTE. Prompts related to different task categories are represented by various colors, while different LLMs are indicated by distinct shapes. For example, a blue square indicates the MTE of Llama2-13B guided by an SA prompt.}
	\label{fig:4} 
\end{figure}

Figure~\ref{fig:4} illustrates the visualization of learned MTEs for LLMs utilizing diverse prompts. 
This visualization reveals two principal clusters formed based on the prompts, highlighting the significant role prompts play in differentiating LLMs performance across task categories. 
While prompts across various tasks significantly influence MTEs, the distinction becomes less pronounced within a single task category, where specific LLMs have a more substantial impact on the embedding outcomes.
Especially within the NLI task category, MTEs are further divided into roughly three clusters, each corresponding to different LLMs.
Mixtral MTEs also generally cluster towards the centre.
See Appendix~\ref{app:vis_llm} for a more detailed visualization, where prompts also slightly influence MTEs within the same task.

\subsubsection{Efficiency Analysis}
\texttt{FUTE} extends the existing task embedding method and utilizes embedding similarity to select prompts, leading to a distinctive computational complexity profile.
The process requires evaluating a total of $k_p$ prompts across $k_D$ datasets.
The computational complexity of \texttt{FUTE} is represented as $O((k_p + k_D)c_o)$, where $c_o$ represents the computation cost associated with the task embedding method.
In contrast, the computational complexity for other baselines evaluated in this study is expressed as $O(k_p k_D c_b)$, where $c_o$ represents the computation cost associated with baselines.
The difference in complexity becomes significant with increasing values of $k_p$ and $k_D$, where the product $k_p k_D$ will exceed the sum $(k_p + k_D)$.
Thus, \texttt{FUTE} presents a potentially more efficient approach to prompt selection, especially as the number of prompts and datasets grows.
However, the existing task embedding methods used in \texttt{FUTE} require fine-tuning, thereby $c_o > c_b$. With a relatively small number of prompts ($k_p$) and datasets ($k_D$), \texttt{FUTE} does not demonstrate superior running efficiency in comparison to baselines.
Nonetheless, $c_o$ is more closely related to the inherent efficiency of the task embedding method itself, whereas \texttt{FUTE} focuses on extending these methods.

\section{Conclusion and Future Work}

This paper introduces a unified framework \texttt{FUTE} that separate task embeddings into DTE and MTE, by employing an additional surrogate base model and unsupervised data, 
without altering existing task embedding learning methodologies. 
\texttt{FUTE} can effectively extend current methods to learn MTEs, which are positioned in a unified vector space, for various language models and LLMs with distinct prompts.
\texttt{FUTE} is, theoretically, capable of learning MTEs for any model with accessible outputs, within a single vector space.
We leave the exploration 
to other models in the future.

\section*{Limitations}

\texttt{FUTE} is designed to extend, rather than enhance, current task embedding learning methods.
Consequently, it inherits potential limitations from the methodologies it builds on.
However, we argue that this is more of a feature of \texttt{FUTE} rather than a limitation. By not altering current procedures, our framework leverages the strengths of existing task embedding methods, allowing for seamless integration of future advancements in the field. This adaptability means that as stronger task embedding methods emerge in the future, our framework will also exhibit improved performance without requiring modifications.
We also acknowledge that any alteration to the surrogate base model requires the re-learning of task embeddings, which could be computationally intensive.
However, we consider this issue intrinsic to the existing task embedding learning methods. 
If future developments in task embedding learning methods eliminate dependency on the base model, \texttt{FUTE} can be adjusted by simply omitting the surrogate base model as well.

\section*{Ethics Statement}
\texttt{FUTE} aims to address challenges related to the interpretability of task embeddings across diverse models, especially for prompt-guided LLMs. While this framework holds promise for enhancing the utility and comparability of task embeddings across different models, it also raises some ethical concerns. For example, learning task embeddings requires the use of unsupervised data, which may not adequately encapsulate the essential attributes of a particular task, potentially resulting in biased task embeddings. Moreover, the reliance on a surrogate T5 model introduces potential ethical issue related to transparency. If the surrogate model fails to accurately replicate the behavior of the target model, it could lead to misleading results.

\section*{Acknowledgements}
This work was supported in part by the UK Engineering and Physical Sciences Research Council (EPSRC) through a Turing AI Fellowship (grant no. EP/V020579/1, EP/V020579/2) and a New Horizons grant (grant no. EP/X019063/1).

\bibliography{custom}

\clearpage
\appendix

\setcounter{figure}{0}
\renewcommand{\thefigure}{A\arabic{figure}}

\setcounter{table}{0}
\renewcommand{\thetable}{A\arabic{table}}

\section{Appendix}
\label{sec:appendix}

\subsection{Datasets in Language Model Experiment}
\label{app:dataset_lm}

Text classification or regression (CR) contain: SNLI \citep{bowman-etal-2015-large}, MNLI \citep{williams-etal-2018-broad}, QQP \citep{quora17}, QNLI \citep{wang2018glue}, SST-2 \citep{socher-etal-2013-recursive}, SciTail \citep{Khot_Sabharwal_Clark_2018}, CoLA \citep{warstadt-etal-2019-neural}, STS-B \citep{cer-etal-2017-semeval}, MRPC \citep{dolan-brockett-2005-automatically}, RTE \citep{dagan2006}, WNLI \citep{Levesque2011TheWS}.

Question answering (QA) datasets contain: SQuAD-2 \citep{rajpurkar-etal-2018-know}, NewsQA \citep{trischler-etal-2017-newsqa}, HotpotQA \citep{yang-etal-2018-hotpotqa}, SQuAD-1 \citep{rajpurkar-etal-2016-squad}, DuoRC-p \citep{saha-etal-2018-duorc}, DuoRC-s \citep{saha-etal-2018-duorc}, DROP \citep{dua-etal-2019-drop}, WikiHop \citep{welbl-etal-2018-constructing}, BoolQ \citep{clark-etal-2019-boolq}, ComQA \citep{abujabal-etal-2019-comqa}, CQ \citep{bao-etal-2016-constraint}.

\subsection{Visualization of Language Model Task Embedding}
\label{app:vis_lm}

\begin{figure*}[t]
	\centering 
	\centerline{\includegraphics[width=0.999\textwidth]{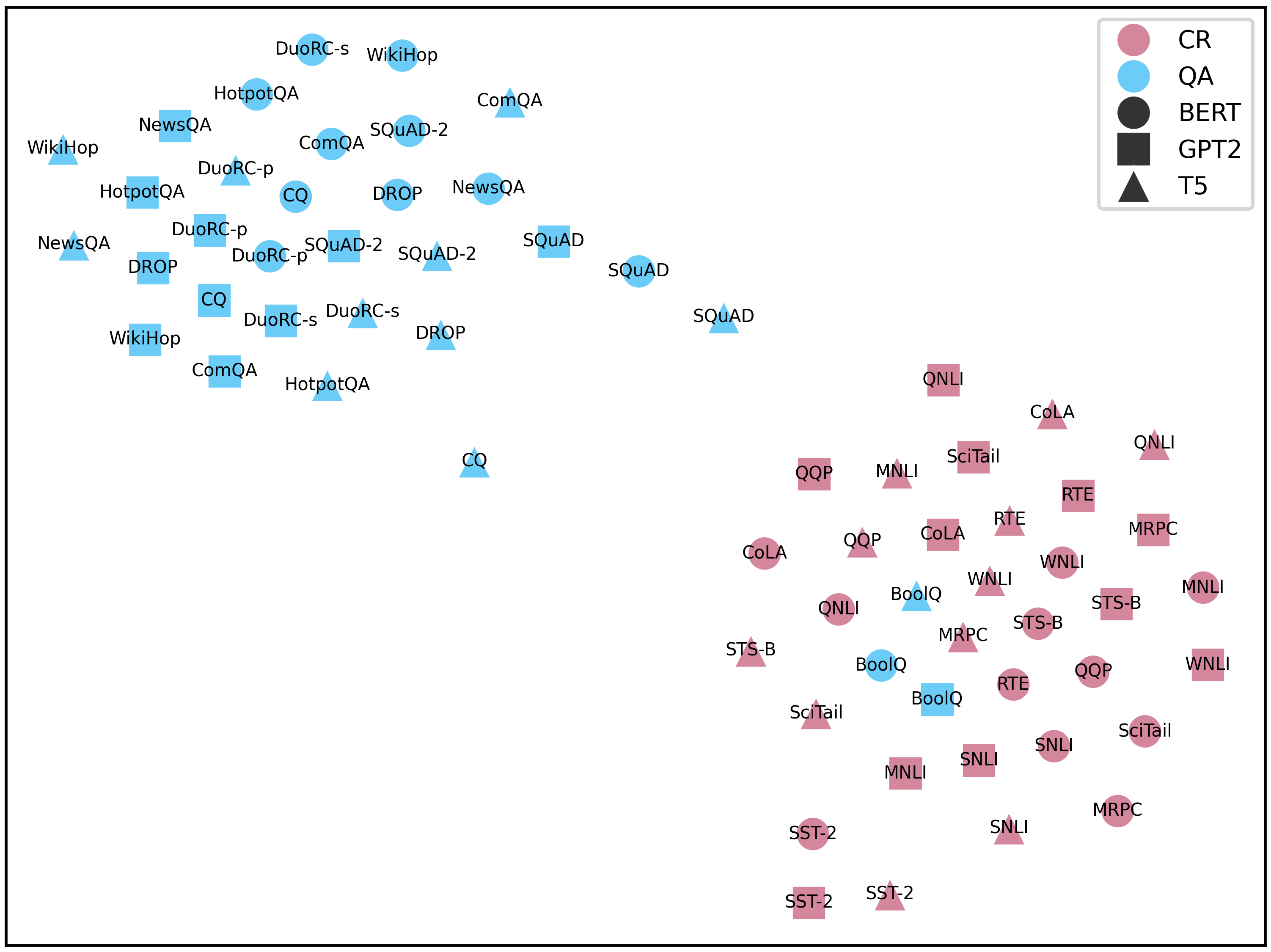}}
	\caption{T-SNE visualization of language model MTEs using \texttt{FUTE} with TuPaTE. Different task categories are represented by various colors, while different models or datasets are indicated by distinct shapes.}
	\label{fig:3a} 
\end{figure*}

Figure~\ref{fig:3a} presents the visualization of task embeddings for language models learned by \texttt{FUTE} with TuPaTE.
This visualization highlights the emergence of distinct clusters that reflect task-specific information, revealing that task embeddings are influenced by both language models and trained datasets:

\paragraph{Datasets} 
Beyond the primary task clusters, some datasets demonstrate a pronounced impact on the formation of task embeddings.
For example, models trained on SNLI and SST-2 exhibit a tendency to cluster together within the same task.
A particular observation is the grouping of three blue points, representing task embeddings associated with the BoolQ, within the CR task cluster. 
This matches the format of BoolQ, a boolean answer for a question, which aligns more closely with CR tasks.

\paragraph{Language Model} 
MTEs associated with the same language model, particularly BERT and GPT2, demonstrate a propensity to cluster together.
This phenomenon indicates that the inherent characteristics of a language model also influence the task embedding space.

\subsection{Prompt Template}
\label{app:instructions}

Table~\ref{table:prompt_sa} and Table~\ref{table:prompt_nli} list used prompts in SA and NLI in Section~\ref{sec:llmexp}. Each contains 10 vanilla prompts and 3 CoT prompts.

\begin{table*}[t]
\centering
\resizebox{0.85\textwidth}{!}{
\begin{tabular}{ll}
\toprule
No. & Prompt \\
\midrule 
1 & Does the movie review below make someone want to watch it? \{text\} \\
2 & \{text\} How does the viewer feel about the movie? \\
3 & \{text\} Did the reviewer enjoy the movie? \\
4 & Does it seem like the reviewer who wrote this review liked the movie? \{text\} \\
5 & Is the movie review below positive? \{text\} \\
6 & \{text\} Did the reviewer find this movie good or bad? \\
7 & \{text\} Is this review positive or negative? \\
8 & \{text\} This is definitely not a positive or negative review \\
9 & \{text\} Did the reviewer enjoy the movie? \\
10 & \{text\} Did the reviewer find this movie good or bad? \\
11 & Does the movie review below make someone want to watch it? \{text\} \\ 
& \quad Reason step by step before answering. \\
12 & \{text\} How does the viewer feel about the movie? \\ 
& \quad Reason step by step before answering. \\
13 & \{text\} Did the reviewer enjoy the movie? \\ 
& \quad Reason step by step before answering. \\
\bottomrule
\end{tabular}
}
\caption{Prompts used for SA}
\label{table:prompt_sa}
\end{table*}

\begin{table*}[t]
\centering
\resizebox{0.99\textwidth}{!}{
\begin{tabular}{ll}
\toprule
No. & Prompt \\
\midrule 
1 & Given that \{premise\} Does it follow that {{hypothesis}} yes, no, or maybe? \\
2 & \{premise\} Using only the above description and what you know about the world, \\
& \quad "\{hypothesis\}" is definitely correct, incorrect, or uncertain? \\ 
3 & \{premise\} Question: \{hypothesis\} true, false, or maybe? \\
4 & Take the following as truth: {{premise}} Then the following statement: \"{{hypothesis}}\" is true, false, or maybe? \\
5 & \{premise\} Question: Does this imply that "\{hypothesis\}"? Yes, no, or maybe? \\
6 & Given \{premise\} Should we assume that "\{hypothesis\}" is true? Yes, no, or maybe? \\
7 & \{premise\} Based on the previous passage, is it true that "\{hypothesis\}"? Yes, no, or maybe? \\
8 & \{premise\} Are we justified in saying that "\{hypothesis\}"? yes, no, or maybe? \\
9 & Suppose it's true that \{premise\} Then, is "\{hypothesis\}" always, sometimes, or never true? \\
10 & Given \{premise\} Is it guaranteed true that "\{hypothesis\}"? yes, no, or maybe? \\
11 & Given that \{premise\} Does it follow that {{hypothesis}} yes, no, or maybe? \\
& \quad Reason step by step before answering. \\
12 & \{premise\} Using only the above description and what you know about the world, \\
& \quad "\{hypothesis\}" is definitely correct, incorrect, or uncertain? \\ 
& \quad Reason step by step before answering. \\
13 & \{premise\} Question: \{hypothesis\} true, false, or maybe? \\
& \quad Reason step by step before answering. \\
\bottomrule
\end{tabular}
}
\caption{Prompts used for NLI}
\label{table:prompt_nli}
\end{table*}

\subsection{Visualization of LLM Task Embedding with Prompts}
\label{app:vis_llm}

\begin{figure*}[t]
	\centering 
	\centerline{\includegraphics[width=0.999\textwidth]{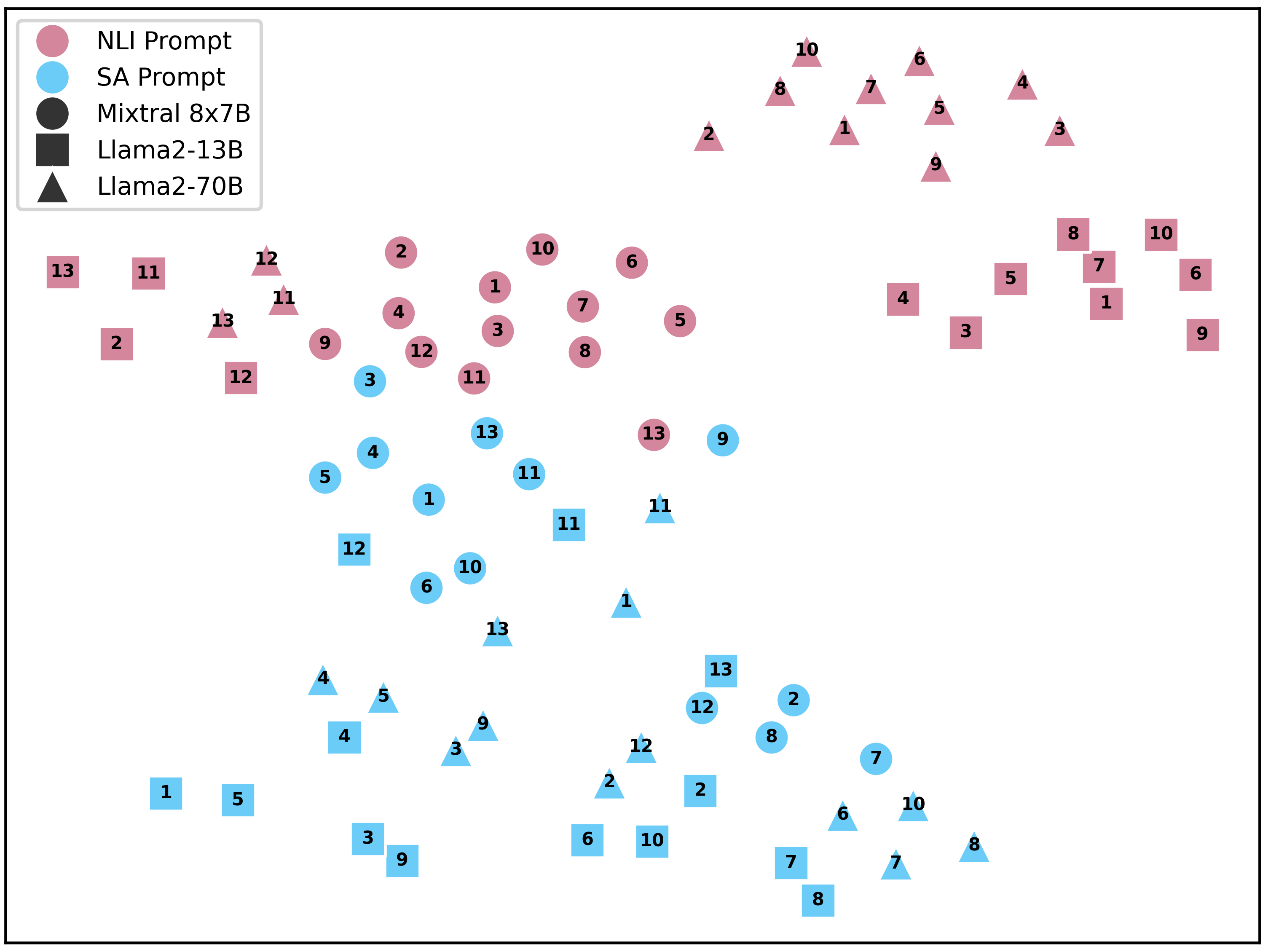}}
	\caption{T-SNE visualization of LLMs MTE using \texttt{FUTE} with TuPaTE. Prompts in different task categories are represented by colors, while different LLMs are indicated by distinct shapes. The numbers correspond to the prompt number as listed in Table~\ref{table:prompt_sa} and Table~\ref{table:prompt_nli}.}
	\label{fig:4a} 
\end{figure*}

Figure~\ref{fig:4a} demonstrates a visualization where prompt numbers are detailed in Table~\ref{table:prompt_sa} and Table~\ref{table:prompt_nli}. Prompt task categories are differentiated by colors, and distinct shapes represent various LLMs.
This visualization reveals that task embeddings are influenced by both the LLMs and the prompts used as follows:

\paragraph{Influence of Prompt}
This visualization demonstrates two influences: inter-task and intra-task influences.
For inter-task influences, task embeddings are clearly separated into two main categories based on the task, indicating that prompts have a substantial impact on LLM performance across different tasks.
For intra-task influences, within the same task, prompts also slightly influence task embeddings.
For example, in the NLI task, all MTEs of prompt\#9 are closer to the SA cluster than MTEs from the same LLM.
Conversely, MTEs of prompt\#10 diverge from the SA cluster.
Also, the Chain of Thought (CoT) prompts\#11, 12, and 13 form their own cluster, distinct from the main LLM-based clusters.
A similar phenomenon is observed within the SA task, where prompts\#7 and 8 diverge from the NLI cluster.

\paragraph{Influence of LLM} 
Task embeddings tend to cluster by LLM, particularly for the NLI task, where embeddings from the same LLM cluster together. 
Mixtral's task embeddings also generally cluster towards the center.

\end{document}